%% file: conference_101719.tex
\documentclass[10pt,conference]{IEEEtran}
\IEEEoverridecommandlockouts
\usepackage{cite}
\usepackage{amsmath,amssymb,amsfonts}
\usepackage{algorithmic}
\usepackage{graphicx}
\usepackage{textcomp}
\usepackage{xcolor}

\usepackage{enumitem}
\usepackage{amsmath,amssymb,mathtools,bm}
\usepackage{graphicx}
\usepackage{booktabs}
\usepackage{enumitem}
\usepackage{multirow} 
\usepackage{svg}
\usepackage[ruled,vlined,linesnumbered]{algorithm2e}
\SetKwInput{KwIn}{Input}
\SetKwInput{KwOut}{Output}
\SetKwRepeat{Repeat}{repeat}{until}
\SetAlgoLined

\makeatletter
\g@addto@macro\normalsize{%
  \setlength\abovedisplayskip{4pt plus 1pt minus 1pt}%
  \setlength\belowdisplayskip{4pt plus 1pt minus 1pt}%
  \setlength\abovedisplayshortskip{2pt plus 1pt minus 1pt}%
  \setlength\belowdisplayshortskip{2pt plus 1pt minus 1pt}%
}
\makeatother

\usepackage{tikz}
\usetikzlibrary{arrows.meta, positioning, fit, calc, shapes.misc}

\def\BibTeX{{\rm B\kern-.05em{\sc i\kern-.025em b}\kern-.08em
    T\kern-.1667em\lower.7ex\hbox{E}\kern-.125emX}}
\begin{document}

\title{FedCVU: Federated Learning for Cross-View \\Video Understanding
}

% \author{
% \IEEEauthorblockN{
% Shenghan Zhang\IEEEauthorrefmark{1}\IEEEauthorrefmark{*} \qquad
% Run Ling\IEEEauthorrefmark{1}\IEEEauthorrefmark{*} \qquad
% Ke Cao\IEEEauthorrefmark{2} \qquad
% Ao Ma\IEEEauthorrefmark{3}\IEEEauthorrefmark{\dagger} \qquad
% Zhanjie Zhang\IEEEauthorrefmark{4}\IEEEauthorrefmark{\dagger}
% }
% \vspace{0.5em}
% \IEEEauthorblockA{
% \IEEEauthorrefmark{1}Software College, Northeastern University, Shenyang, China \\
% \IEEEauthorrefmark{2}University of Science and Technology of China, Hefei, China \\
% \IEEEauthorrefmark{3}University of Chinese Academy of Sciences, Beijing, China \\
% \IEEEauthorrefmark{4}Zhejiang University, Hangzhou, China \\
% \IEEEauthorrefmark{*}Equal contribution \\
% \IEEEauthorrefmark{\dagger}Corresponding authors: maaoaoma@126.com, cszzj@zju.edu.cn \\
% zhangsh24@mails.neu.edu.cn, runling001@hotmail.com
% }
% }

\author{
\IEEEauthorblockN{
Shenghan Zhang\textsuperscript{1,*} \qquad
Run Ling\textsuperscript{1,*} \qquad
Ke Cao\textsuperscript{2} \qquad
Ao Ma\textsuperscript{3,$\dagger$} \qquad
Zhanjie Zhang\textsuperscript{4,$\dagger$}
}
\vspace{0.5em}
\IEEEauthorblockA{
\textsuperscript{1}Software College, Northeastern University, Shenyang, China \\
\textsuperscript{2}University of Science and Technology of China, Hefei, China \\
\textsuperscript{3}University of Chinese Academy of Sciences, Beijing, China \\
\textsuperscript{4}Zhejiang University, Hangzhou, China \\
% \textsuperscript{*}Equal contribution \\
% \textsuperscript{$\dagger$}Corresponding author \\
zhangsh24@mails.neu.edu.cn, runling001@hotmail.com
}
}

\maketitle

\begin{abstract}
Federated learning (FL) has emerged as a promising paradigm for privacy-preserving multi-camera video understanding. 
However, applying FL to cross-view scenarios faces three major challenges: (i) heterogeneous viewpoints and backgrounds lead to highly non-IID client distributions and overfitting to view-specific patterns, (ii) local distribution biases cause misaligned representations that hinder consistent cross-view semantics, and (iii) large video architectures incur prohibitive communication overhead. 
To address these issues, we propose FedCVU, a federated framework with three components: VS-Norm, which preserves normalization parameters to handle view-specific statistics; CV-Align, a lightweight contrastive regularization module to improve cross-view representation alignment; and SLA, a selective layer aggregation strategy that reduces communication without sacrificing accuracy. 
Extensive experiments on action understanding and person re-identification tasks under a cross-view protocol demonstrate that FedCVU consistently boosts unseen-view accuracy while maintaining strong seen-view performance, outperforming state-of-the-art FL baselines and showing robustness to domain heterogeneity and communication constraints.

% Cross-view video understanding with federated learning (FL) is challenged by strong view-induced heterogeneity and the high communication cost of large video models. Under these constraints, synchronizing all network layers across clients is neither efficient nor always beneficial, as indiscriminate aggregation can amplify view-specific variations and compromise generalization to unseen views. This raises a central problem in cross-view FL: how to decide which network layers should be globally synchronized and which should remain local under limited communication budgets.

% We propose FedCVU, a federated framework that addresses this problem through transferability-aware selective layer aggregation. We characterize layer-wise transferability as whether aggregating a layer’s updates across clients leads to consistent global improvement under view variations, and introduce a lightweight estimator based on cross-client update agreement and effective update strength to guide synchronization. To support reliable estimation, FedCVU incorporates view-specific normalization and a prototype-based contrastive alignment to stabilize semantic representations.
% Experiments on cross-view action recognition and video-based person re-identification show that FedCVU improves unseen-view performance while reducing communication cost by nearly half compared with strong FL baselines.

\end{abstract}


\section{Introduction}
% 背景
Federated learning (FL) has emerged as a promising paradigm for privacy-preserving collaborative model training without centralizing raw data~\cite{mcmahan2017communication}, enabling applications in sensitive domains such as healthcare~\cite{rieke2020future}, finance~\cite{yang2019federated}, and computer vision~\cite{chen2021bridging}.
% FL 在视频理解中的应用
In video understanding, FL is particularly appealing for multi-camera systems where centralizing footage is often infeasible due to privacy regulations or bandwidth constraints~\cite{yang2025privacy, tu2024fedfslar}. By allowing each camera or site to train locally while sharing only model updates, FL mitigates the risk of data leakage and enables large-scale collaboration across distributed video streams.

% FL 在视频理解中面临的挑战
% However, applying FL to cross-view video understanding faces three major challenges.
% First, multi-camera video streams exhibit large variations in viewpoint, background, illumination, and scene composition~\cite{rahmani2017learning, liu2018viewpoint}. Such heterogeneity leads to highly non-IID client distributions that impede optimization and cause models to overfit to view-specific patterns~\cite{li2020federated, karimireddy2020scaffold}.
% Second, even with shared model updates, the representations learned by different clients remain misaligned due to local distribution biases, making it difficult to capture consistent cross-view semantics~\cite{li2018unsupervised, wu2019cross}.
% Third, modern video architectures such as 3D convolutional networks~\cite{tran2015learning} and vision transformers~\cite{dosovitskiy2020image, arnab2021vivit} contain tens to hundreds of millions of parameters, leading to prohibitive communication overhead in each FL round~\cite{konevcny2016federated}.
% These challenges pose significant obstacles to deploying FL-based solutions in real-world multi-camera systems.

However, applying FL to cross-view video understanding faces three major challenges.
First, multi-camera video streams exhibit large variations in viewpoint, background, illumination, and scene composition~\cite{rahmani2017learning, liu2018viewpoint}. Such variations are not random but are tightly coupled with camera geometry and deployment environments, resulting in highly structured and persistent feature shifts across clients. This form of heterogeneity leads to highly non-IID client distributions that impede optimization and cause models to overfit to view-specific patterns~\cite{li2020federated, karimireddy2020scaffold}, thereby limiting generalization to unseen camera views.
Second, even with shared model updates, the representations learned by different clients often remain misaligned due to local distribution biases. In cross-view settings, semantically identical actions or identities may exhibit drastically different visual appearances across viewpoints, making it difficult for standard FL aggregation to induce consistent representations.
As a result, the global model may converge to view-dependent features that fail to capture cross-view semantics~\cite{li2018unsupervised, wu2019cross}. Third, modern video architectures such as 3D convolutional networks~\cite{tran2015learning} and vision transformers~\cite{dosovitskiy2020image, arnab2021vivit} contain tens to hundreds of millions of parameters. Directly synchronizing such large models across many clients incurs prohibitive communication overhead in each FL round~\cite{konevcny2016federated}, which is further exacerbated by the slow convergence typically observed under severe non-IID conditions.
These challenges are closely intertwined, posing obstacles to deploying FL-based solutions for real-world multi-camera video understanding systems.

% 方法
To address these challenges, we propose FedCVU, a federated framework for cross-view video understanding. 
FedCVU integrates three key components. 
\emph{View-Specific Normalization (VS-Norm)} preserves client-specific normalization parameters while aggregating the rest, reducing view-dependent distribution gaps and enhancing generalization to unseen cameras. 
\emph{Cross-View Contrastive Alignment (CV-Align)} introduces a lightweight contrastive loss that encourages consistent representations across clients, mitigating semantic misalignment. 
\emph{Selective Layer Aggregation (SLA)} dynamically selects layers for aggregation under a communication budget, substantially lowering overhead while maintaining accuracy. 
Together, these modules jointly address domain heterogeneity, representation inconsistency, and communication efficiency, providing a principled solution for multi-camera video understanding.

% 实验
We evaluate FedCVU under a cross-view protocol on two representative tasks: action understanding and person re-identification. 
Across both datasets, FedCVU achieves consistent gains on unseen-view accuracy while preserving strong performance on seen views. 
Compared with competitive federated baselines, it further attains higher accuracy with substantially reduced communication cost, confirming its robustness to domain heterogeneity, representation misalignment, and bandwidth constraints.
Our main contributions are summarized as follows:
\begin{itemize}
    \item We introduce FedCVU, a federated framework for cross-view video understanding that addresses the key challenges of domain heterogeneity, representation misalignment, and communication overhead.  
    \item We develop three dedicated modules: VS-Norm for mitigating view-specific distribution gaps, CV-Align for enforcing cross-view representation alignment, and SLA for reducing communication cost through selective aggregation.
    \item We conduct comprehensive experiments on action understanding and person re-identification tasks under a cross-view evaluation protocol, demonstrating consistent gains in unseen-view accuracy with competitive seen-view performance over strong FL baselines.  
\end{itemize}

\section{Related Works}

\subsection{Federated Learning}
Federated Learning (FL) enables collaborative training without sharing raw data \cite{mcmahan2017communication}, but suffers from performance degradation under non-IID data. 
Prior work improves robustness through optimization strategies such as proximal regularization \cite{li2020federated}, variance reduction \cite{karimireddy2020scaffold}, dynamic regularization \cite{acar2021federated}, and adaptive server updates \cite{reddi2020adaptive}. 
These methods mainly stabilize training across heterogeneous clients, yet overlook structured feature shifts caused by different acquisition conditions.
Personalization methods further address heterogeneity. FedBN \cite{li2021fedbn} retains local normalization statistics, while MOON \cite{li2021model} introduces contrastive objectives for representation consistency. 
However, most approaches are designed for image tasks and do not explicitly model cross-view semantic alignment in federated video settings. 
Our work focuses on view-induced heterogeneity and representation misalignment in federated video understanding.

\subsection{Cross-View Video Understanding}
Cross-view video understanding has been widely studied in centralized settings, especially for action recognition and person re-identification. 
Existing methods learn view-invariant representations via metric learning or feature alignment \cite{rahmani2017learning,liu2018viewpoint,wu2019cross}, supported by benchmarks such as MCAD \cite{li2017multi} and MARS \cite{zheng2016mars}.
These approaches rely on centralized multi-view data and joint optimization across cameras, which is incompatible with federated settings where data are isolated. 
Consequently, independently trained client models may produce misaligned representations after aggregation. 
Our method explicitly models cross-view consistency under federated constraints to improve generalization to unseen views.

\section{Method}

\subsection{Problem Formulation}  
% We consider $C$ clients collaboratively training a global video understanding model without sharing raw data. Client $c$ holds $\mathcal{D}^c=\{(x_j^c, y_j^c)\}_{j=1}^{n_c}$, where $x_j^c\in\mathbb{R}^{3\times T\times H\times W}$ is a video and $y_j^c\in\{1,\dots,K\}$ is its label. Due to video variations, the underlying distribution $P_c(x,y)$ differs across clients, yielding a highly non-IID setting. The objective is
% \begin{equation}
% \min_{\theta} \sum_{c=1}^{C} \frac{n_c}{N}\, \mathcal{L}_c(\theta), 
% \mathcal{L}_c(\theta) = \mathbb{E}_{(x,y)\sim \mathcal{D}^c} \big[-\log p_\theta(y|x)\big],
% \end{equation}
% where $N = \sum_{c=1}^C n_c$. We adopt a \emph{cross-view} protocol: training on \emph{seen views} and evaluating on \emph{unseen views}, aiming to improve generalization to unseen viewpoints while retaining competitive performance on seen ones.

% In this section, we present FedCVU, a federated learning framework tailored for cross-view video understanding, followed by its enhanced variant FedCVU-BN, which preserves local batch normalization statistics to reduce viewpoint-specific gaps. 
% We further incorporate two modules, lightweight contrastive regularization (LCR) and optional layer-wise aggregation (OLA), to improve cross-view feature consistency and communication efficiency. 
% The overall training process is illustrated in Figure~\ref{fig:method} and summarized in Algorithm~\ref{alg:fedcvu}.

% \textbf{Problem Formulation.}  
We study cross-view federated video understanding, where $C$ clients collaboratively train a global video classification model without sharing raw footage. 
Client $c$ holds a local dataset $\mathcal{D}^c=\{(x_j^c, y_j^c)\}_{j=1}^{n_c}$, where $x_j^c\in\mathbb{R}^{3\times T\times H\times W}$ is a video clip and $y_j^c\in\{1,\dots,K\}$ the class label. 
Due to differences in viewpoint, background, and illumination, the local data distribution $P_c(x,y)$ varies significantly across clients, leading to a highly non-IID setting that is challenging for standard FL algorithms. 
Formally, the global optimization objective is:
\begin{align}
\min_{\theta} \sum_{c=1}^{C} \frac{n_c}{N} \, \mathcal{L}_c(\theta), \quad N = \sum_{c=1}^C n_c,
\end{align}
where the local classification loss is
\begin{equation}
\mathcal{L}_c(\theta) =
\mathbb{E}_{(x,y)\sim \mathcal{D}^c}
\left[ -\log p_\theta(y|x) \right].
\end{equation}
We adopt a cross-view evaluation protocol in which training is conducted on videos from a subset of viewpoints (\emph{seen views}) and testing on held-out viewpoints (\emph{unseen views}). 
The main objective is to maximize unseen-view accuracy while maintaining competitive seen-view performance.

\subsection{Federated Framework: FedCVU}

% We propose FedCVU, a federated framework for cross-view video understanding. It integrates three designs: VS-Norm for mitigating view-specific distribution gaps, CV-Align for aligning cross-view representations, and SLA for reducing communication cost.

We propose FedCVU, a unified federated framework that jointly considers view heterogeneity, representation consistency, and communication efficiency.
Instead of handling these factors independently, FedCVU integrates three complementary components that interact within a single training loop.
View-Specific Normalization stabilizes local feature statistics across camera views, Cross-View Contrastive Alignment promotes semantic consistency among client representations, and Selective Layer Aggregation prioritizes communication on transferable model components.
Together, these designs enable effective cross-view generalization under federated constraints.

\subsubsection{View-Specific Normalization}
A key source of heterogeneity in cross-view federated learning arises from distribution shifts in feature statistics across camera viewpoints.
Differences in camera placement, illumination conditions, and scene composition induce systematic variations in feature activations, which are directly reflected in normalization statistics.
Under standard federated aggregation, averaging all parameters, including normalization layers, forces clients to share global statistics that often misrepresent their local distributions.
This mismatch can distort feature scaling and bias intermediate representations toward dominant views, ultimately harming cross-view generalization.

To address this issue, we propose View-Specific Normalization (VS-Norm), which preserves client-specific normalization parameters while aggregating the rest of the model.
By decoupling normalization from global synchronization, VS-Norm enables each client to maintain feature statistics that are well aligned with its local camera view, while still participating in collaborative representation learning.

Formally, let $\theta=\{\theta_{\text{norm}},\theta_{\text{rest}}\}$ denote the global parameters, where $\theta_{\text{norm}}$ includes the normalization-specific weights.
For CNN-based backbones with \emph{Batch Normalization (BN)}, we have $\theta_{\text{BN}}=\{\gamma,\beta,\mu,\sigma\}$ representing scale, shift, and running statistics.
In FedCVU, only $\theta_{\text{rest}}$ is aggregated, while each client $c$ maintains its own normalization parameters $\theta_{\text{BN}}^c$:
\begin{equation}
\text{BN}^c(x) = \gamma^c \cdot \frac{x - \mu^c}{\sqrt{(\sigma^c)^2 + \epsilon}} + \beta^c,
\end{equation}
where $\mu^c$ and $\sigma^c$ are estimated from client $c$’s local data.

For Transformer-based backbones with \emph{Layer Normalization (LN)}, $\theta_{\text{LN}}=\{\gamma,\beta\}$ denotes the affine scale and shift, and each client retains its own copy $\theta_{\text{LN}}^c$.
This design generalizes VS-Norm across different video architectures without introducing additional parameters or communication cost.
By preserving view-specific normalization while sharing higher-level representations, VS-Norm reduces harmful statistical interference across clients and provides a stable foundation for subsequent cross-view representation alignment.

\subsubsection{Cross-View Contrastive Alignment}

Even with shared updates, client models often learn misaligned representations due to local distribution biases, making cross-view semantics inconsistent.
In cross-view scenarios, samples belonging to the same semantic class may exhibit large appearance variations across camera views, which can cause the global model to drift toward view-dependent features.
To alleviate this issue, we introduce Cross-View Contrastive Alignment (CV-Align), a lightweight contrastive regularization scheme that encourages semantic consistency across clients while preserving local training autonomy.

Concretely, given a mini-batch $\{(x_i, y_i)\}$ on client $c$, we extract representations $h_i = f_\theta(x_i) \in \mathbb{R}^d$.
Each class $y$ is associated with a global prototype $z_y \in \mathbb{R}^d$, which summarizes the mean embedding of class $y$ aggregated across clients and communication rounds.
These prototypes serve as a shared semantic anchor, enabling indirect alignment among clients without requiring direct data sharing.

During training, prototypes are maintained on the server and updated via exponential moving average (EMA):
\begin{equation}
z_y \leftarrow \mu z_y + (1-\mu) \cdot \frac{1}{|\mathcal{B}_y|} \sum_{i: y_i=y} h_i ,
\end{equation}
where $\mathcal{B}_y$ denotes the set of samples with label $y$ in the current round, and $\mu \in [0,1)$ controls the update momentum.
This update scheme allows prototypes to evolve smoothly over rounds, providing a stable yet adaptive reference for cross-view alignment.

The alignment loss is defined as a contrastive objective between each local embedding $h_i$ and the prototype set $\{z_y\}$:
\begin{equation}
\mathcal{L}_{\text{CV-Align}} = - \sum_{i} \log 
\frac{\exp(\text{sim}(h_i, z_{y_i}) / \tau)}
     {\sum_{y'=1}^K \exp(\text{sim}(h_i, z_{y'}) / \tau)} ,
\end{equation}
where $\text{sim}(\cdot,\cdot)$ denotes cosine similarity and $\tau$ is a temperature parameter.
By pulling embeddings toward their corresponding class prototypes while pushing them away from others, CV-Align promotes view-invariant semantic representations across clients.

Finally, the client’s training objective is formulated as
\begin{equation}
\mathcal{L}_c = \mathcal{L}_{\text{CE}} + \mathcal{L}_{\text{CV-Align}},
\end{equation}
where $\mathcal{L}_{\text{CE}}$ is the standard cross-entropy loss.
In this way, CV-Align acts as an auxiliary regularizer that complements the primary classification objective without introducing additional model parameters or communication overhead.

% \subsubsection{Cross-View Contrastive Alignment.} 
% Even with shared updates, client models often learn misaligned representations due to local distribution biases, making cross-view semantics inconsistent. 
% To address this, we introduce CV-Align, a lightweight contrastive regularization scheme. 

% Concretely, given a mini-batch $\{(x_i, y_i)\}$ on client $c$, we extract representations $h_i = f_\theta(x_i) \in \mathbb{R}^d$. 
% Each class $y$ is associated with a global prototype $z_y \in \mathbb{R}^d$, which summarizes the mean embedding of class $y$ aggregated across clients and rounds. 
% During training, prototypes are updated on the server by exponential moving average (EMA):
% \begin{equation}
% z_y \leftarrow \mu z_y + (1-\mu) \cdot \frac{1}{|\mathcal{B}_y|} \sum_{i: y_i=y} h_i ,
% \end{equation}
% where $\mathcal{B}_y$ is the set of samples with label $y$ in the current round, and $\mu \in [0,1)$ is the momentum coefficient.

% The alignment loss is then defined as a contrastive objective between each local embedding $h_i$ and the prototype set $\{z_y\}$:
% \begin{equation}
% \mathcal{L}_{\text{CV-Align}} = - \sum_{i} \log 
% \frac{\exp(\text{sim}(h_i, z_{y_i}) / \tau)}
%      {\sum_{y'=1}^K \exp(\text{sim}(h_i, z_{y'}) / \tau)} ,
% \end{equation}
% where $\text{sim}(\cdot,\cdot)$ denotes cosine similarity and $\tau$ is a temperature parameter. 
% Finally, the client’s training objective becomes
% \begin{equation}
% \mathcal{L}_c = \mathcal{L}_{\text{CE}} + \mathcal{L}_{\text{CV-Align}},
% \end{equation}
% where $\mathcal{L}_{\text{CE}}$ is the standard cross-entropy loss.

\begin{algorithm}[th!]
\caption{Federated Learning with FedCVU}
\label{alg:fedcvu}
\KwIn{Clients $\{1,\dots,C\}$ with data $\{\mathcal{D}^c\}$; rounds $T$; local epochs $E$; SLA budget $B$; decision interval $\tau$.}
\KwOut{Global model $\theta$ and prototypes $\{z_y\}$.}

\For{round $t=1$ \KwTo $T$}{
  \tcp{Server $\rightarrow$ Clients}
  Broadcast global model $\theta^{(t)}$ and prototypes $\{z_y\}$\;

  \tcp{Clients (in parallel)}
  \For{each client $c$}{
    \textbf{VS-Norm:} keep local norm parameters frozen (others trainable)\;
    \textbf{Local training:} for $E$ epochs, minimize ($\mathcal{L}_{\text{CE}}$ + $\mathcal{L}_{\text{CV-Align}}$) using $\{z_y\}$\;
    \textbf{Layer signatures:} for each block $\ell$, compute and keep
      normalized gradient direction $\hat g_\ell^{c,(t)}$ and norm $r_\ell^{c,(t)}$\;
    Upload updated model $\theta^{c,(t)}$ (with local norm kept local), and signatures
      $\{\hat g_\ell^{c,(t)}, r_\ell^{c,(t)}\}_\ell$ to server\;
  }

  \tcp{Server: SLA statistics \& selection}
  Aggregate client signatures to obtain per-block agreement $\kappa_\ell^{(t)}$ and salience $s_\ell^{(t)}$\;
  Compute per-block utility $u_\ell^{(t)}$ and cost $b_\ell$; if $t \bmod \tau = 0$, select subset
    $\mathcal{S}_t$ under budget $B$ via greedy on $u_\ell^{(t)}/b_\ell$ (always include blocks 1,2 and L$-$1,L)\;

  \tcp{Server: soft weights \& thresholded gating}
  For each block $\ell$, set soft weight $w_\ell^{(t)}$:
    $w_\ell^{(t)}{=}1$ if $\ell\!\in\!\mathcal{S}_t$; otherwise use a small weak-sync weight in $(0,\lambda]$\;
  If $w_\ell^{(t)} < \eta$, gate it out this round (no parameter upload/download for $\ell$; signatures only), and set $\tilde w_\ell^{(t)}{=}0$; else $\tilde w_\ell^{(t)}{=}w_\ell^{(t)}$\;

  \tcp{Server: weighted aggregation}
  Update global blocks: 
  $\theta_\ell^{(t+1)} \leftarrow (1-\tilde w_\ell^{(t)})\,\theta_\ell^{(t)} + \tilde w_\ell^{(t)} \cdot \sum_c \frac{n_c}{N}\,\theta_\ell^{c,(t)}$ for $\ell$\;

  \tcp{Server: prototype maintenance}
  Update prototypes $\{z_y\}$ via EMA using client embeddings/statistics from round $t$\;
}
\end{algorithm}

\input{Tables/performance_comparison}

\subsubsection{Selective Layer Aggregation}

To alleviate the prohibitive communication cost of large video models, we propose Selective Layer Aggregation (SLA), which selectively synchronizes only the most informative and transferable layers across clients.
In cross-view settings, different layers exhibit varying degrees of view sensitivity: shallow and deep layers tend to encode more transferable low-level cues and high-level semantics, while intermediate layers are often strongly affected by viewpoint-specific patterns.
Uniform synchronization therefore leads to unnecessary communication and may even harm personalization.

Instead of transmitting full gradients, each client $c$ extracts \emph{lightweight layer signatures} for block $\ell$ at round $t$:
\begin{equation}
\hat g_\ell^{c,(t)} \;=\; \frac{g_\ell^{c,(t)}}{\|g_\ell^{c,(t)}\|_2}, 
\end{equation}
\begin{equation}
r_\ell^{c,(t)} \;=\; \|g_\ell^{c,(t)}\|_2,
\end{equation}
where $g_\ell^{c,(t)}$ is the local gradient (or parameter update).
These signatures capture both the update direction and its magnitude with negligible communication overhead.

On the server, signatures from all clients are aggregated to compute two per-block statistics:
\begin{equation}
\kappa_\ell^{(t)} \;=\; \frac{2}{C(C-1)} \sum_{c<c'} 
\big\langle \hat g_\ell^{c,(t)}, \hat g_\ell^{c',(t)} \big\rangle,
\end{equation}
\begin{equation}
s_\ell^{(t)} \;=\; \Big\| \frac{1}{C}\sum_{c=1}^C r_\ell^{c,(t)}\,\hat g_\ell^{c,(t)} \Big\|_2,
\end{equation}
where $\kappa_\ell^{(t)}$ measures cross-client directional agreement, reflecting representation consistency, and 
$s_\ell^{(t)}$ measures the salience of the aggregated update, reflecting its potential impact on global learning.

We define the utility $u_\ell^{(t)}=\max\{0,\kappa_\ell^{(t)}\}\cdot s_\ell^{(t)}$ and the cost $b_\ell$ as the bytes of block $\ell$.
Given a per-round budget $B$, SLA selects a subset $\mathcal S_t$ via a knapsack objective (greedy on $u_\ell^{(t)}/b_\ell$):
\begin{equation}
\mathcal S_t=\arg\max_{\mathcal S} \sum_{\ell\in\mathcal S} u_\ell^{(t)}
\quad\text{s.t.}\quad \sum_{\ell\in\mathcal S} b_\ell \le B.
\end{equation}
To stabilize training, the shallowest (1, 2) and deepest (L$-$1, L) blocks are always synchronized, and the selection is recomputed every $\tau$ rounds.

We adopt two modes of block aggregation:
\textbf{Strong Synchronization} ($\ell\in\mathcal S_t$) with fixed weight $1$, and 
\textbf{Weak Synchronization} ($\ell\notin\mathcal S_t$) with a capped soft weight:
\begin{equation}
w_\ell^{(t)}=
\begin{cases}
1, & \ell\in\mathcal S_t, \\[3pt]
\lambda\,\sigma\!\big(\alpha(\kappa_\ell^{(t)}-\tau_\kappa)\big), & \ell\notin\mathcal S_t,
\end{cases}
\end{equation}
where $\sigma$ is the sigmoid, $\tau_\kappa$ the agreement threshold, and $\lambda\!\le\!0.3$ limits weak synchronization.

\emph{Thresholded gating:} if $w_\ell^{(t)}<\eta$, we drop parameter upload/download for block $\ell$ in round $t$ (signatures only), and set the effective weight $\tilde w_\ell^{(t)}{=}0$; otherwise $\tilde w_\ell^{(t)}{=}w_\ell^{(t)}$.
The global update is
\begin{equation}
\theta_{\ell}^{(t+1)} \;=\; (1-\tilde w_\ell^{(t)})\,\theta_{\ell}^{(t)} \;+\; \tilde w_\ell^{(t)}\!\!\sum_{c}\tfrac{n_c}{N}\,\theta_{\ell}^{c,(t)}.
\end{equation}
Overall, SLA prioritizes blocks that are both consistent across clients and cost-effective to communicate, while allowing view-sensitive layers to remain personalized, achieving an effective balance between communication efficiency and cross-view generalization.

\textbf{Overall Algorithm.}
The complete training procedure of FedCVU is summarized in Algorithm~\ref{alg:fedcvu}.
It follows the standard federated loop of local training and server aggregation, while integrating VS-Norm, CV-Align, and SLA within a unified optimization process.
Specifically, VS-Norm stabilizes local feature statistics during client-side training, CV-Align enforces cross-view semantic consistency through prototype-based regularization, and SLA adaptively controls parameter synchronization under a communication budget.
Together, these components enable efficient and robust federated learning for cross-view video understanding.

\section{Experiments}
\subsection{Experimental Setup}

\textbf{Datasets.}
We evaluate FedCVU on two benchmarks: MCAD~\cite{li2017multi} for action understanding and MARS~\cite{zheng2016mars} for video person re-identification. 
MCAD contains multi-view surveillance videos, while MARS consists of pedestrian tracklets from 6 cameras . 
We follow a cross-view protocol, training on \emph{seen} cameras and testing on \emph{unseen} ones. 
To simulate federated clients, each dataset is split into 20 clients by evenly dividing cameras. 
Unlike prior works that employ synthetic Dirichlet splits, we adopt this camera-based partition to better reflect real-world deployment. 
% Videos are sampled into 16-frame clips, resized to 224$\times$224.

\textbf{Baselines.} 
We compare with \emph{FedAvg}~\cite{mcmahan2017communication}, 
\emph{FedProx}~\cite{li2020federated}, 
\emph{SCAFFOLD}~\cite{karimireddy2020scaffold}, 
\emph{MOON}~\cite{li2021model}, 
\emph{FedBN}~\cite{li2021fedbn}, 
\emph{FedDyn}~\cite{acar2021federated}, 
and \emph{FedOpt}~\cite{reddi2020adaptive}, 
covering proximal regularization, variance reduction, contrastive learning, BN personalization, dynamic regularization, and adaptive server optimization. 

\textbf{Implementation Details.} 
A pretrained 3D VAE encoder extracts frozen video latents. 
The federated model is a Transformer ($d{=}512$, $L{=}12$, $f{=}4$, $\sim$37.8M params) with a lightweight classification head. 
Only Transformer and head parameters are aggregated; VS-Norm stats remain local. 
We train with AdamW (lr=1e-4, weight decay=0.05, cosine decay, 5 local epochs) and use BF16 precision for communication. 
Metrics are Top-1/Top-5 (MCAD) and mAP/CMC@1 (MARS). 
Experiments run on 8$\times$A100 GPUs, and results are averaged over three seeds.

% \subsection{Overall Performance Comparison.}
% Table~\ref{tab:main_results} shows that FedCVU consistently achieves the best unseen-view performance on both MCAD and MARS. It reaches 83.1\% Top-1 on MCAD and 73.2\% mAP on MARS, outperforming strong baselines such as FedBN and FedOpt by clear margins. While simpler methods like FedAvg and FedProx converge in fewer rounds, they saturate at much lower accuracy, highlighting their limited generalization under cross-view shifts. In contrast, stronger baselines achieve better final accuracy but require more rounds.
% As shown in Fig.~\ref{fig:convergence}, FedCVU provides the best trade-off: it converges smoothly within 112–123 rounds and reaches the highest accuracy while reducing communication cost by nearly half. These results validate the effectiveness of our design in balancing accuracy, robustness, and efficiency for cross-view video understanding.

\subsection{Overall Performance Comparison}

Table~\ref{tab:main_results} reports the performance of FedCVU and competing methods on MCAD and MARS under the cross-view protocol with 20 clients.
FedCVU consistently achieves the best unseen-view performance on both tasks, reaching 83.1\% Top-1 accuracy on MCAD and 73.2\% mAP on MARS.
Compared with strong baselines such as FedBN and FedOpt, FedCVU improves Top-1 accuracy by 1.8--2.3\% on MCAD and mAP by 2.5--3.0\% on MARS, indicating more effective cross-view generalization.
Simple aggregation-based methods such as FedAvg and FedProx converge in fewer communication rounds, but their performance saturates at substantially lower accuracy.
This gap highlights their limited ability to cope with severe view-induced non-IID distributions.
Methods designed to stabilize federated optimization or improve representation consistency, including SCAFFOLD, MOON, FedBN, FedDyn, and FedOpt, achieve markedly higher accuracy, yet still fall short of FedCVU, especially on unseen views.
These results suggest that addressing optimization or normalization alone is insufficient for robust cross-view video understanding.

\begin{figure}[t!]
    \centering
    \includegraphics[width=0.5\textwidth]{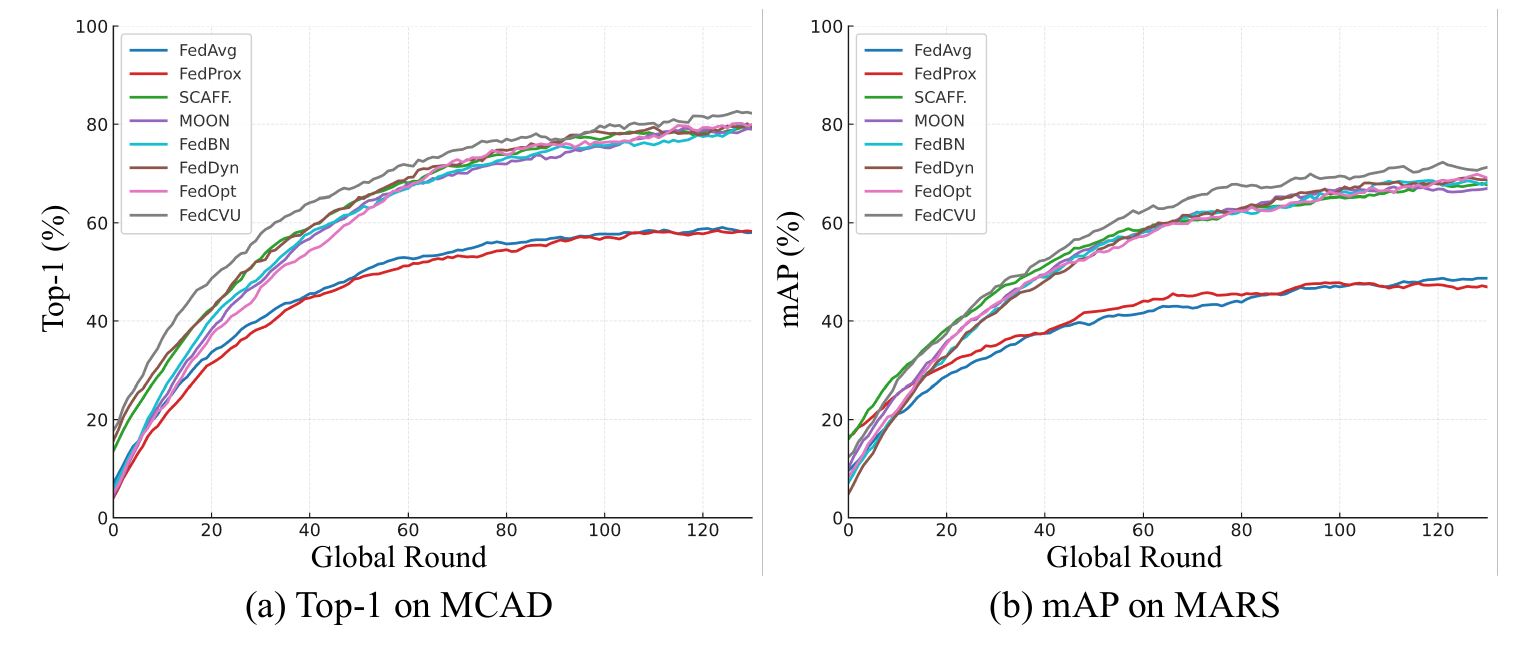}
    % \vspace{-1.0em}
\caption{
Convergence curves on MCAD (Top-1) and MARS (mAP). FedCVU converges smoothly within 112–123 rounds and achieves the highest accuracy, while FedAvg/FedProx plateau early at low performance.
}
    \label{fig:convergence}
    % \vspace{-1.0em}
\end{figure}

In terms of efficiency, FedCVU also demonstrates a clear advantage.
Despite achieving the highest accuracy, it reduces per-client communication cost to 5.8~GB on MCAD and 6.0~GB on MARS, corresponding to nearly a 40--45\% reduction compared with standard FL baselines.
Moreover, FedCVU converges within 112--123 rounds, which is comparable to or faster than most strong baselines.
As shown in Fig.~\ref{fig:convergence}, FedCVU exhibits smooth and stable convergence behavior, achieving a favorable balance between accuracy, robustness, and communication efficiency.
% Overall, these results validate the effectiveness of FedCVU for cross-view federated video understanding across both action recognition and person re-identification tasks.

\input{Tables/ablation_mcad}
\input{Tables/ablation_mars}

\subsection{Ablation Studies}

We analyze the contribution of each module in FedCVU on MCAD and MARS, as shown in Tables~\ref{tab:ablation_mcad} and~\ref{tab:ablation_mars}.
Overall, removing any component leads to consistent performance degradation, indicating that the three modules are complementary rather than redundant.

On MCAD, removing VS-Norm results in a clear drop of Top-1 accuracy ($-1.6\%$) and Top-5 accuracy ($-1.1\%$), highlighting its effectiveness in mitigating view-specific distribution gaps induced by different camera viewpoints.
Eliminating CV-Align also degrades performance (Top-1 $-0.8\%$, Top-5 $-0.7\%$), suggesting that representation alignment further improves cross-view generalization beyond statistical normalization alone.
These results indicate that both local feature stabilization and cross-client semantic alignment are important for action understanding under cross-view federated settings.

On MARS, the impact of removing VS-Norm or CV-Align is even more pronounced.
Discarding either module leads to a decrease of $1.4$–$1.6\%$ in mAP and around $0.3\%$ in CMC@1, reflecting the sensitivity of person re-identification to view-dependent appearance variations.
This observation confirms that maintaining consistent identity representations across camera views benefits from the combined effects of view-specific normalization and prototype-based alignment.

In contrast, removing SLA yields only marginal accuracy degradation on both datasets (less than $0.3\%$), while substantially increasing communication cost by more than $50\%$.
This demonstrates that SLA primarily improves communication efficiency without compromising predictive performance.
Together, these ablation results show that VS-Norm and CV-Align play a critical role in improving accuracy and robustness, whereas SLA effectively reduces communication overhead, enabling FedCVU to achieve a favorable balance between performance and efficiency.

\begin{figure}[htp!]
    \centering
    \includegraphics[width=0.45\textwidth]{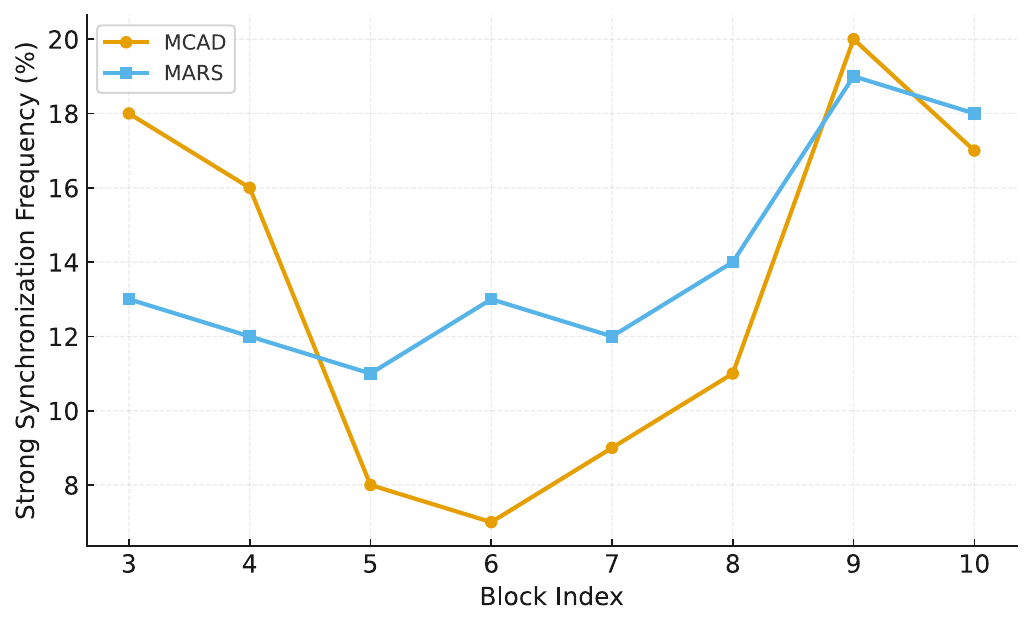}
    % \vspace{-1.0em}
\caption{Strong synchronization frequency across Transformer blocks on MCAD and MARS. 
MCAD exhibits a U-shaped pattern where shallow and deep blocks are more frequently synchronized, while mid-level blocks remain mostly localized due to higher view-specific variability. 
In contrast, MARS shows consistently higher synchronization in deeper blocks, reflecting the stability of identity semantics across cameras.}
    \label{fig:sla_strong_sync_line}
    % \vspace{-1.0em}
\end{figure}

\subsection{Analysis of Synchronization Frequency} 
Fig.~\ref{fig:sla_strong_sync_line} reports the proportion of rounds in which each block is strongly synchronized under SLA. 
On MCAD, the curve follows a U-shape: shallow blocks (3--4) and deep blocks (9--10) are frequently synchronized, while mid-level blocks (5--7) are less often selected. 
This pattern is consistent with the nature of cross-view action understanding, where low-level motion cues and high-level semantics are more transferable across views, whereas mid-level features are highly view-dependent and thus better preserved locally. 
In contrast, MARS shows a steadily increasing trend toward deeper blocks, with Blocks 9--10 dominating synchronization. 
This reflects the fact that person re-identification primarily relies on identity-level representations, which are more consistent across clients and therefore favored by SLA. 
Together, these results confirm that SLA adaptively concentrates communication on the most globally consistent layers while allowing view-specific layers to remain personalized.

\section{Conclusion}
\label{sec:conclustion}
We presented FedCVU, a federated framework for cross-view video understanding that jointly addresses view-induced heterogeneity, representation misalignment, and communication efficiency.
FedCVU integrates view-specific normalization, cross-view contrastive alignment, and selective layer aggregation within a unified training process.
% Extensive experiments demonstrate consistent improvements in generalization to unseen camera views.
% Future work will explore extensions to multi-modal settings across views.

\bibliographystyle{IEEEtran}
\bibliography{refs}

\clearpage
\newpage
\appendix
\section{Connection to Modern Representation Learning and Generative Models}

\subsection{Relation to Contrastive Representation Learning}

Recent advances in contrastive learning, including SimCLR~\cite{chen2020simpleframeworkcontrastivelearning}, MoCo~\cite{he2020momentumcontrastunsupervisedvisual}, and CLIP~\cite{radford2021learningtransferablevisualmodels}, have demonstrated the effectiveness of learning transferable representations by maximizing agreement between semantically related samples while separating unrelated ones. These methods typically rely on large-scale centralized data and instance-level or cross-modal alignment, enabling strong generalization across domains and tasks.

Beyond instance-level contrastive objectives, extensions to supervised and prototype-based settings~\cite{snell2017prototypicalnetworksfewshotlearning} further improve semantic consistency by explicitly leveraging class-level structure. Such approaches replace pairwise matching with alignment to class centers or prototypes, providing a more stable and semantically meaningful learning signal, especially under intra-class variation.

In federated learning, contrastive methods such as MOON~\cite{li2021modelcontrastivefederatedlearning} have been introduced to alleviate client drift by enforcing consistency between local and global representations. However, existing approaches mainly focus on model-level alignment and do not explicitly address structured feature shifts caused by heterogeneous data distributions, such as those induced by different camera views in video scenarios.

Our CV-Align builds upon these lines of work by introducing a prototype-based contrastive mechanism tailored for cross-view federated settings. Instead of relying on shared data or direct sample matching, CV-Align maintains global class prototypes as lightweight semantic anchors, enabling consistent alignment across clients with disjoint and heterogeneous data. This design bridges supervised contrastive learning and federated optimization, and is particularly suited for cross-view video understanding where semantically identical patterns may exhibit substantial appearance variation across viewpoints.

\subsection{Relation to Video and Image Generation Models}

Recent progress in image and video generation has been largely driven by diffusion models and diffusion transformers, which enable high-fidelity synthesis through iterative denoising and powerful latent representations~\cite{ho2020denoising, rombach2022high}. These models have been extended to video generation, where additional challenges such as temporal consistency, multi-subject interaction, and long-range coherence arise.

To address these challenges, a number of recent works have explored structured control and representation learning within generative frameworks. For example, methods such as FancyVideo~\cite{fancyvideo} and MoFu~\cite{mofu} improve temporal consistency and multi-subject generation through cross-frame guidance and frequency-aware fusion. WISA~\cite{wisa} introduces physics-aware simulation priors to enhance realism, while Video-EM~\cite{videoem} focuses on long-form video understanding via persistent memory mechanisms. Efficiency-oriented designs, such as Qihoo-T2X~\cite{qihoo_t2x} and RelaCtrl~\cite{relactrl}, further improve scalability by reducing redundant computation or introducing relevance-guided control. In the image domain, works like RAGAR~\cite{ragar} and U-StyDiT~\cite{ustydit} explore personalization and high-quality style transfer within diffusion transformer frameworks.

Despite these advances, most existing methods assume centralized training with access to large-scale datasets, and rely on shared representations or explicit conditioning to achieve cross-view or multi-condition consistency. In contrast, real-world multi-camera or multi-source scenarios often involve distributed and privacy-sensitive data, where centralized training is infeasible.

Our FedCVU framework provides a complementary perspective by addressing representation consistency under federated constraints. In particular, the proposed CV-Align introduces prototype-based alignment to enforce cross-view semantic consistency without sharing raw data, while VS-Norm stabilizes view-specific feature distributions. These properties are directly relevant to generative modeling, where maintaining consistent identity, motion, and semantics across views or conditions is critical. Therefore, FedCVU can be viewed as a step toward federated generative modeling, offering a potential pathway to extend diffusion-based video and image generation to distributed settings with heterogeneous data sources.

\subsection{Generalization to Generative and Multi-Modal Settings}

Although FedCVU is evaluated on discriminative tasks such as action understanding and person re-identification, its core design is not limited to classification settings. The proposed framework addresses a more general problem: learning consistent and transferable representations under heterogeneous and distributed data distributions.

In generative modeling, especially in video and image generation, maintaining semantic consistency across conditions, views, or modalities is a fundamental challenge. For example, multi-view video generation requires preserving identity and motion coherence across viewpoints, while text-to-video models must align visual content with semantic conditions over time. These requirements are closely related to the cross-view representation alignment problem studied in this work.

The components of FedCVU can be naturally extended to such scenarios. VS-Norm provides a mechanism to handle condition-specific or domain-specific feature shifts, which are common in multi-modal or multi-view generation. CV-Align offers a lightweight way to enforce semantic consistency without requiring shared data, making it suitable for distributed generative training. Meanwhile, SLA enables efficient communication by prioritizing globally consistent and transferable components, which is particularly important for large-scale generative models with high communication cost.

These properties suggest that FedCVU can serve as a general framework for federated representation learning beyond discriminative tasks. In particular, it provides a promising direction for extending diffusion-based generative models and multi-modal learning systems to federated settings, where data are inherently decentralized and heterogeneous.
% \bibliographystyle{IEEEtran}
% \bibliography{refs}

\end{document}

%% file: Tables/performance_comparison.tex
\begin{table*}[t]
\centering
\caption{
Results on MCAD (action understanding) and MARS (person re-identification) under the cross-view protocol with 20 clients. 
Mean $\pm$ std over three runs is reported. 
\textbf{Bold} and \underline{underline} mark best and second-best mean values; scores within one std are regarded as comparable. 
Comm. (GB) is the per-client communication cost (BF16, upload+download) until convergence; $R^{*}$ is the average number of rounds to converge.
}
\label{tab:main_results}
\resizebox{\linewidth}{!}{
\begin{tabular}{lccccccccc}
\toprule[1.1pt]
\textbf{Metric} & \textbf{Oracle} & \textbf{FedAvg} & \textbf{FedProx} & \textbf{SCAFF.} & \textbf{MOON} & \textbf{FedBN} & \textbf{FedDyn} & \textbf{FedOpt} & \textbf{FedCVU} \\
\midrule
\multicolumn{10}{l}{\textit{MCAD (Action Understanding)}} \\
\midrule
Top-1 (\%)      & \textbf{84.2} & 58.9$\pm$0.3 & 58.6$\pm$0.4 & 80.4$\pm$0.5 & 80.1$\pm$0.4 & \underline{81.3$\pm$0.3} & 81.0$\pm$0.5 & 80.8$\pm$0.4 & \textbf{83.1$\pm$0.2} \\
Top-5 (\%)      & \textbf{96.5} & 60.8$\pm$0.1 & 61.2$\pm$0.2 & 91.6$\pm$0.3 & 91.4$\pm$0.2 & \underline{92.1$\pm$0.2} & 91.9$\pm$0.2 & 91.7$\pm$0.2 & \textbf{94.3$\pm$0.2} \\
Comm. (GB)  & -- & 9.7$\pm$0.5 & 9.4$\pm$0.4 & 8.8$\pm$0.4 & 8.6$\pm$0.3 & 8.5$\pm$0.3 & 8.3$\pm$0.2 & 8.2$\pm$0.2 & 5.8$\pm$0.1 \\
$R^{*}$ (rounds)& --   & 95$\pm$4   % FedAvg (差 → 早收敛)
& 97$\pm$4   % FedProx (差 → 早收敛)
& 110$\pm$5  % SCAFF. (中等)
& 112$\pm$4  % MOON (中等)
& 118$\pm$4  % FedBN (较好 → 慢一些)
& 116$\pm$3  % FedDyn (中等靠上)
& 120$\pm$3  % FedOpt (强)
& 112$\pm$3 \\
\midrule
\multicolumn{10}{l}{\textit{MARS (Person Re-Identification)}} \\
\midrule
mAP (\%)        & \textbf{77.4} & 48.1$\pm$0.5 & 47.9$\pm$0.4 & 69.3$\pm$0.3 & 69.8$\pm$0.3 & 70.5$\pm$0.4 & \underline{70.7$\pm$0.3} & 70.2$\pm$0.4 & \textbf{73.2$\pm$0.3} \\
CMC@1 (\%)      & \textbf{89.7} & 63.0$\pm$0.4 & 63.7$\pm$0.3 & 84.1$\pm$0.3 & 84.8$\pm$0.2 & 85.1$\pm$0.3 & \underline{85.3$\pm$0.3} & 85.0$\pm$0.2 & \textbf{87.4$\pm$0.2} \\
Comm. (GB)  & -- & 10.2$\pm$0.5 & 9.9$\pm$0.4 & 9.3$\pm$0.4 & 9.1$\pm$0.3 & 9.0$\pm$0.3 & 8.8$\pm$0.2 & 8.6$\pm$0.2 & 6.0$\pm$0.1 \\
$R^{*}$ (rounds)& --   
& 102$\pm$4   % FedAvg
& 104$\pm$4   % FedProx
& 118$\pm$5   % SCAFF.
& 120$\pm$4   % MOON
& 126$\pm$4   % FedBN
& 124$\pm$3   % FedDyn
& 129$\pm$3   % FedOpt
& 123$\pm$3 \\
\bottomrule[1.1pt]
\end{tabular}}
\end{table*}

%% file: Tables/ablation_mcad.tex
\begin{table}[t]
\centering
\caption{Ablation study of FedCVU on MCAD (action understanding). 
Mean $\pm$ std over three runs.}
\vspace{0.1em}
\label{tab:ablation_mcad}
\resizebox{\linewidth}{!}{
\begin{tabular}{lccc}
\toprule[1.1pt]
\textbf{Variant} & Top-1 (\%) & Top-5 (\%) & Comm. (GB) \\
\midrule
w/o VS-Norm   & 81.5$\pm$0.3 & 93.2$\pm$0.3 & 5.9$\pm$0.2 \\
w/o CV-Align  & 82.3$\pm$0.3 & 93.6$\pm$0.3 & 5.7$\pm$0.1 \\
w/o SLA       & 82.9$\pm$0.2 & 94.0$\pm$0.2 & 8.8$\pm$0.3 \\
\textbf{FedCVU (Full)} & \textbf{83.1$\pm$0.2} & \textbf{94.3$\pm$0.2} & \textbf{5.8$\pm$0.1} \\
\bottomrule[1.1pt]
\end{tabular}}
\end{table}

%% file: Tables/ablation_mars.tex
\begin{table}[t]
\centering
\caption{Ablation study of FedCVU on MARS (person re-identification). 
Mean $\pm$ std over three runs.}
\vspace{0.1em}
\label{tab:ablation_mars}
\resizebox{\linewidth}{!}{
\begin{tabular}{lccc}
\toprule[1.1pt]
\textbf{Variant} & mAP (\%) & CMC@1 (\%) & Comm. (GB) \\
\midrule
w/o VS-Norm   & 71.8$\pm$0.3 & 86.1$\pm$0.3 & 6.1$\pm$0.1 \\
w/o CV-Align  & 71.6$\pm$0.3 & 86.0$\pm$0.3 & 6.0$\pm$0.1 \\
w/o SLA       & 72.9$\pm$0.3 & 87.1$\pm$0.2 & 9.3$\pm$0.3 \\
\textbf{FedCVU (Full)} & \textbf{73.2$\pm$0.3} & \textbf{87.4$\pm$0.2} & \textbf{6.0$\pm$0.1} \\
\bottomrule[1.1pt]
\end{tabular}}
\end{table}